\documentclass{article}
\usepackage{ijcai21}

\usepackage{times}

\usepackage{soul}
\usepackage{url}
\usepackage[hidelinks]{hyperref}
\usepackage[utf8]{inputenc}
\usepackage[small]{caption}
\usepackage{graphicx}
\usepackage{eqnarray,amsmath}
\usepackage{booktabs}
\usepackage{amssymb}
\usepackage{algorithm}
\usepackage[noend]{algpseudocode}

\usepackage{subfig}

\makeatletter
\newcommand{\thickhline}{%
    \noalign {\ifnum 0=`}\fi \hrule height 1pt
    \futurelet \reserved@a \@xhline
}
\newcolumntype{"}{@{\hskip\tabcolsep\vrule width 1pt\hskip\tabcolsep}}
\makeatother

\title{A distillation-based approach integrating continual learning and federated learning for pervasive services}

\author{
Anastasiia Usmanova$^1$\footnote{This work has been partially supported by the LabEx PERSYVAL-Lab (ANR-11-LABX-0025-01) and by MIAI@Grenoble Alpes (ANR-19-P3IA-0003) funded by the French program Investissement d’avenir.}\and
François Portet$^2$\and
Philippe Lalanda $^2$
and German Vega$^2$\\
\affiliations
$^1$Grenoble INP, France\\
$^2$Univ. Grenoble Alpes, CNRS, Inria, Grenoble INP, LIG, 38000 Grenoble, France\\
\emails
\ anastasiia.usmanova@grenoble-inp.org,
Firstname.name@imag.fr
}

\begin{document}

\maketitle

\begin{abstract}
Federated Learning, a new machine learning paradigm enhancing the use of edge devices, is receiving a lot of attention in the pervasive community to support the development of smart services. Nevertheless, this approach still needs to be adapted to the speciﬁcity of the pervasive domain. In particular, issues related to continual learning need to be addressed. In this paper, we present a distillation-based approach dealing with catastrophic forgetting in federated learning scenario. Specifically, Human Activity Recognition tasks are used as a demonstration domain.
\end{abstract}

\section{Introduction}
\vspace{-0.1cm}

Pervasive computing promotes the integration of connected devices in our living spaces in order to assist us in our daily activities.
The devices collect the data, can run some local computations and further give advises to users, act upon the environment through the services or just provide information to a global server.

We are now seeing the emergence of smarter services based on Machine Learning (ML) techniques \cite{becker}. Moreover, current implementations are actually based on distributed architectures where models are deployed and often executed in the cloud. This approach is, however, not well adapted to pervasive computing. It undergoes major limitations in terms of security, performance and cost. 

% \smallskip

% One interesting solution is to make more advanced use of edge resources like gateways or smartphones. The notion of the edge was mentioned in 2009 \cite{2009}  and generalized by Cisco Systems in 2014 as a new operational model. The main idea is to place computing and storage functions as close as possible to data sources. Regarding machine learning, it comes down to ofﬂoading some learning tasks in edge resources. In doing so, data to be transferred can be reduced, and reactivity can be improved. Also, privacy can be better preserved since the nature of the exchanged information is modiﬁed.

Google recently proposed Federated Learning (FL) \cite{mcmahan} \cite{bonawitx}, a new Machine Learning paradigm enhancing the use of edge devices. FL encourages the computation of local models on edge devices and sending them to a cloud server where they are aggregated into a more generic one. The new model is redistributed to devices as a bootstrap model for the next local learning iteration. FL reduces communication costs and improves security because only models are exchanged between the edge and cloud \cite{8}. It has immediately attracted attention as a promising paradigm that can meet the challenges of ML-based pervasive applications. Nevertheless, this approach still needs to be adapted to the speciﬁcity of the pervasive domain. 
% In that domain, models are more likely to diverge because of data and clients heterogeneity, and because of different data volumes generated by clients

In most current solutions, FL operates with static local data which stays the same during the whole training process for each client \cite{percom}. However, in real world scenarios, new data on edges is continuously available and models have to adapt to it while not forgetting the past experience. This requires the use of \textit{Continual Learning} (CL) approaches \cite{cl}. The main challenge of CL  with deep neural networks is \textit{catastrophic forgetting} \cite{catastrophic-forgetting}, \cite{ewc}, which is caused by optimizing the entire network according to the new tasks and not anymore to older tasks. 
% This is the consequence of a more general neural network's problem known as the \textit{stability-plasticity} dilemma \cite{stab-plas}, where plasticity and stability refer to integrating new knowledge and retaining previous knowledge respectively. 

CL is characterized by sequential nature of the learning process, concretely by sequences of \textit{tasks}. \textit{Task $t$} is a set of classes disjoint from classes in other (previous or future) tasks, where $t$ is its task-ID \cite{class-inc}.
%It means that data sequentially arrives in batches which corresponds to one task that can be defined as a set of classes which are disjoint from classes in other tasks \cite{class-inc}.
Mostly CL methods address  \textit{task-incremental learning} (task-IL)
scenario \cite{task-inc}, where information about the task-ID of examples is known at test time. However, more challenging scenario is \textit{class-incremental learning} (class-IL), where the model has to distinguish among all the classes of all the tasks at  test time \cite{class-inc}. 

In our  work, we focus on a class-IL scenario of CL, which is combined with FL scenario. Our purpose was to tackle the following research questions: does FL help to prevent catastrophic forgetting on a client side? can FL help to share the past knowledge of all clients and improve their performance on unknown tasks? 
%what CL approaches can be helpful in FL scenario? and 
can we take an advantage of global server to improve the performance of clients?   

We decided to use Human Activity Recognition classification (HAR) on mobile devices as our demonstration domain \cite{lara,blachon:hal-01082580} due to  availability of “natural” clients: smartphones, and that this domain is not that well investigated as image classification. This choice is challenging due to the close relation of some classes (movement actions).

% We believe that HAR is an excellent domain for our investigation. We appreciate the availability of diverse datasets, diverse approaches with good performances that can be used for comparison, the imbalanced nature of data, and the availability of “natural” clients: smartphones. 

In this paper, we propose a distillation-based method which deals with catastrophic forgetting in Federated Continual Learning (FCL) for the HAR classification task. In section 2, we  give some background on FL and CL fields and describe current existing solutions regarding the FCL problem. In section 3, we present the methodology of our work. In section 4, we propose our method. In section 5, we show experimental results and make a discussion of them. In section 5, we finish the work with a conclusion.

\section{Background. Related work}

\subsection{Federated Learning}

The main goal of conventional FL is to achieve a good global model on a server. FL process includes multiple communication rounds of: 1) local training on a client side (with static local datasets), after that clients send their updated parameters to the server; 2) parameters of gotten models are aggregated by the server to define a new global model which is further redistributed to all the clients to become an initial model in the next communication round.

% and parameters aggregation on a server side. 
% %At each round, each client makes a local training on its local dataset (usually on the same one big dataset).
% After the training each client sends its updated parameters to the server. The server aggregates of all the parameters gotten from clients and define a new global model which is further redistributed to all clients/ New communication round starts. Various aggregation techniques have been defined. 

FedAvg \cite{fedavg} uses a simple weighted average of clients' models where weights are based on the number of examples in a local dataset. FedPer \cite{fedper}  separates layers of clients' models on base and specialization layer and sends to the server only base layers. FedMa \cite{fedma} makes a layer-wise aggregation with the fusion of similar neurons.

In our work, we want to achieve a good performance both on a server and on a clients' side, as they are the main potential users of the models. For this, we will use the FedAvg as a base method. It shows competitive result on HAR \cite{percom} and doesn't require much computational resources that is crucial for mobile devices.

\subsection{Continual Learning}

In standard CL on a single machine, a model iteratively learns from a sequence of tasks. At each training session, the model can have access to only one task from the sequence in order. After the training, the task is not accessible anymore.

For class-IL scenarios, CL methods can be divided into three families \cite{class-inc}: \textit{regularisation-based} approaches which compute an importance of weights for previous tasks and penalise the model for changing them (EWC \cite{ewc}, MAS \cite{mas}, PathInt \cite{pathint}, LwF \cite{lwf}),  \textit{exemplar-based} approaches which store exemplars from previous tasks (iCarl \cite{icarl}, EEIL \cite{eeil}), and \textit{bias-correction} approaches which deal explicitly with bias towards recently-learned tasks (BiC \cite{bic}, LUCIT \cite{lucir}, IL2M \cite{il2m}).

\subsection{Federated Continual Learning}

In FCL each client has its privately accessible sequence of tasks. Each round client trains its local model on some task from its sequence and then sends its parameters o the server.

To the best of our knowledge, only few works based on a fusion of Federated and Continual Learning have been proposed so far.

For the task-IL image classification problem in FL, FedWeIT \cite{fedweit} was recently proposed. It is based on a decomposition of the model parameters into a dense global parameters
%(generic knowledge of all tasks of all clients) 
and sparse task-adaptive 
%(knowledge for concrete client and concrete task),
which are shared between all the clients. 
%This CL implementation (APD \cite{apd}) belongs to parameter-isolation methods, which are not appropriate for the class-IL scenario \citep{3cl}, \citep{class-inc}.
LFedCon2 \cite{fcl-nodnn}  and FedCL \cite{fcl} deals with single-task scenario in FL. LFedCon2 uses traditional classifiers instead of DNN and propose an algorithm dealing with a concept drift based on ensemble retraining. FedCL adopt EWC \cite{ewc} (regularization-based  CL method which does not show the best result for the class-IL scenario \cite{class-inc}). FedCL aims to improve the generalization ability of federated models.

However, none of these approaches solves the problem of class-IL CL scenario  in  FL which  we  investigate  in  our  work. That is why our research has a novel contribution in this field.

\newpage
%%%%%%%%%%%%%%%%%%%%%%%%%%%%%%%%%%%%%%%%%%%%%%%%%%%%%%%%%%
%%%%%%%%%%%%%%%%%%%%%%%%%%%%%%%%%%%%%%%%%%%%%%%%%%%%%%%%%%
%%%%%%%%%%%%%%%%%%%%%%%%%%%%%%%%%%%%%%%%%%%%%%%%%%%%%%%%%%

\section{Methodology}

%In this section, we describe the specificity of HAR and the features of the dataset we use, define our methodology and present a set of experiments. 

\subsection{Human Activity Recognition. UCI dataset}

We work with a UCI HAR dataset \cite{uci} which is widely used by the HAR research community to measure and compare state-of-the-art results. The UCI dataset was collected with a Samsung Galaxy S II placed on the waist with the help of 30 volunteers. One example of data contains 128 recordings of accelerometer and gyroscope (both 3 dimensions). There are 6 classes in the dataset (number of examples per class are in brackets):  0 -- Walking (1722), 1 -- Walking Up (1544), 2 -- Walking Down (1406), 3 -- Sitting (1777), 4 -- Standing (1906), 5 -- Laying (1944).

\begin{figure}[!ht]
  \centering
  \includegraphics[width=\linewidth]{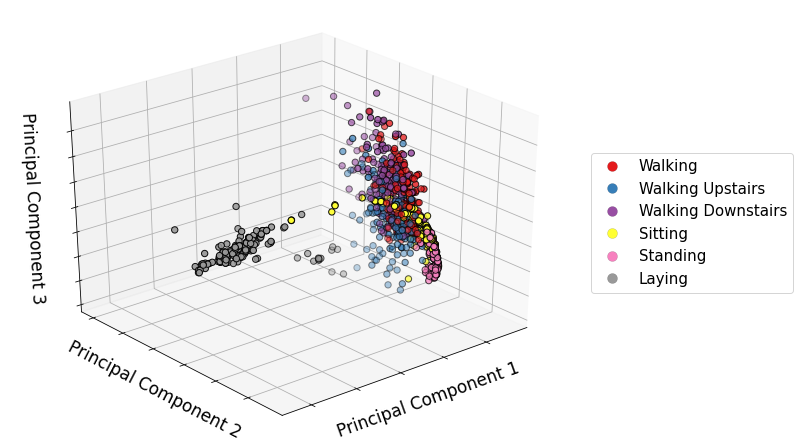}
  \caption{PCA of UCI dataset on a last layer of the used neural net, 3D space of first 3 principal components.}
  \label{fig:pca}
\end{figure}

To understand the relation between classes, we made a Principal Component Analysis (PCA) \cite{pca} of a UCI dataset on a last layer (before the activation) of a neural net used in our experiments (more in Section 3.3). We randomly chose 200 examples of each class for better representation and plot the first 3 principal  components of this data in a 3 dimensional space (see Figure \ref{fig:pca}). 

We can see that the class "Laying" is very distinguishable from the rest. "Sitting" and "Standing" locate  close  to  each  other.  Walking  movements locate  together,  but individually it is hard to distinguish them. Such a data distribution will have an impact on catastrophic forgetting of some classes in our experiments. For example, on Figure \ref{fig:cat-for0}, we will see that class 5 (Laying) is never forgotten even when there is no example of this class during a training session.

\subsection{Assumptions}

\begin{itemize}
  \item We focus on a \textbf{class-incremental CL scenario} since we want to classify among all the seen classes, as more common in real world settings.
   \vspace{-0.15cm}\item \textbf{At each communication round}, each  client  trains  its  own  model  on \textbf{a new  local  dataset} corresponding  to some task, as in real world new data is collected on mobile devices in a time between communication with the server. Data from previous rounds is not available, unless otherwise mentioned.
   \vspace{-0.15cm}\item As we still in our work adhere the definition of a task, \textbf{any task can’t be seen in a private sequence of tasks for each client more than once.} However, sets of classes corresponding to tasks from private sequences of different clients can be overlapped.
   \vspace{-0.15cm}\item For the sake of simplicity, \textbf{clients behave synchronously}, so all of them take part in each communication round.
\end{itemize}

\subsection{FCL problem definition}

Under the assumptions above, we present following notations:

\begin{itemize}
\vspace{-0.2cm} \item each client $k \in \{ 1, 2, ..., K \}$  has its privately accessible sequence of $n_k$ tasks $\mathcal{T}_k$:
\vspace{-0.1cm} 
\begin{equation}
\mathcal{T}_k =  \left[\mathcal{T}_k^{1}, \mathcal{T}_k^{2}, ... , \mathcal{T}_k^{t}, ... , \mathcal{T}_k^{n_k}\right],
\mathcal{T}_k^{t} = (C^t_k, D^t_{k}),  \nonumber
\label{eq:task_seq_fcl}
\end{equation}

\vspace{-0.1cm} 

\noindent where $t \in \{1,...,n_k\}$, $C^t_k$ is a set of classes which represent the task $t$ of a client $k$ and $D^t_k =  \{X^t_k,Y^t_k\}$ is a training data corresponding to $C^t_k$:  $C^{i}_k \cap C^{j}_k= \varnothing$ if $i \neq j$;

\vspace{-0.07cm} \item each task $\mathcal{T}_k^{t}$ for the client $k$ is trained during $r^t_k$ communication rounds and $\sum_{t=1}^{n_k}r^t_k = R$, where $R$ is a total number of communication rounds between the server and clients;
\vspace{-0.07cm} \item during communication round $r$ client $k$ uses training data $D_{kr} = \{X_{kr}, Y_{kr}\}$:

\vspace{-0.6cm} 

\begin{equation}
D_{kr} = D^t_{kr} \subset D^t_k, \hspace{2mm}  \sum_{t=1}^{t-1}r^t_k < r \leqslant \sum_{t=1}^{t-1}r^t_k + r^t_k, \nonumber
\end{equation}

\vspace{-0.25cm} 

\noindent  where $D_{k'r'} \cap D_{k''r''} = \emptyset $, if $k' \neq k''$ and $r' \neq r''$

\noindent ($1 \leqslant k',k'' \leqslant K, 1 \leqslant r',r'' \leqslant R$);
\end{itemize}

\subsection{Clients' scenarios. Train and test sets}

As the main CL scenario in FL, we observe the behaviour of a client which learns one task during half of the total number of communication rounds, and second half it learns a new task. 

\smallskip

In notations above, \textbf{client 1} learns $n_1 = 2$ tasks in total: $\mathcal{T}_1^1 = (C_1^1, D_1^1)$ and $\mathcal{T}_1^2 = (C_1^2, D_1^2)$, where $C^1_1 = \{1\}$ and $C^2_1 = \{2\}$  -- only "Walking Up" and only "Walking Down" classes, respectively; $\mathcal{T}_1 = \left[\mathcal{T}_1^{1}, \mathcal{T}_1^{2}\right]$,  $r_1^1=r_1^2=R/2$.

\smallskip

% \begin{itemize}
% \vspace{-0.05cm}
%  \item learns $n_1 = 2$ tasks in total: $\mathcal{T}_1^1 = (C_1^1, D_1^1)$ and $\mathcal{T}_1^2 = (C_1^2, D_1^2)$, where $C^1_1 = \{1\}$ and $C^2_1 = \{2\}$
%     -- only "Walking Up" and only "Walking Down" classes, respectively;
    
% \item \vspace{-0.1cm} has its privately accessible sequence of tasks $\mathcal{T}_1 = \left[\mathcal{T}_1^{1}, \mathcal{T}_1^{2}\right]$;

%     \item \vspace{-0.1cm} learns both tasks $\mathcal{T}_1^1$ and $\mathcal{T}_1^2$ during $r_1^1=r_1^2=R/2$ communication rounds.
% \end{itemize}

For the simplicity, we assume that all other $K-1$ clients behave similarly, and we can generalize their influence in FL process by the single \textbf{generalized client}. We also assume that all clients contain the same number of examples at each round, so the size of clients dataset will not influence forgetting. That's why, during the server aggregation the weights of generalized and the observed clients are $1/K * (K-1)$ and $1/K$ respectively.

The generalized client performs online-learning (training on the same task each round) on data which is well-balanced and contain all the classes. So, it learns $n_g = 1$ task in total: $\mathcal{T}_g^1 = (C_g^1, D_g^1)$, where $C^1_g = \{0,1,2,3,4,5\}$, has its privately accessible sequence of tasks $\mathcal{T}_g = \left[\mathcal{T}_g^{1}\right]$, so $r_g^1=R$.

\medskip
To build train and test sets we randomly chose examples from UCI HAR dataset according to required CL scenario for each client. As mentioned above, for each client $k$ and for each communication round $r$ we built a \textbf{train set} $D_{kr}$ \textbf{of the same size}.

We estimate the performance of the models on a \textbf{common test set} for all clients and the server. The test dataset includes 100 examples of each class (600 examples in total).

\newpage 
\subsection{Neural Network architecture}

The server and all clients use the same Neural Network architecture. As we should take into account a small processing power on mobile devices, we want to limit complexity and size of a model we use. We made a comparison of different architectures among Convolution Neural Networks (CNN) which were used in other state-of-the-art studies \cite{ignatov}, \cite{wu-cnn}.

\smallskip

By centralized approach, we trained the models on the dataset UCI HAR used for all the experiments in our work: 70\% of data is a train set, 15\% is a validation set, 15\% is a test set. The models are trained using a mini-batch SGD of size 32 and a dropout rate of 0.5. 

\vspace{-0.1cm} 

\begin{table}[!ht]
  \begin{center}
    \begin{tabular}{|c|c|c|} 
      \hline
      \textbf{Model Architecture} & \textbf{Model Size} & \textbf{Test Acc} \\
      \hline
      32-10C\_256D\_128D & 3.3 MB & 92.67\\
      64-9C\_4M\_32-9C\_2M\_16-9C\_128D & 45 KB & 93.76 \\
      \textbf{196-16C\_4M\_1024D} & 67.8 MB & \textbf{94.64}\\
      196-16C\_4M\_1024D\_512D &  74.1 MB & 92.27\\
      \hline
    \end{tabular}
    \caption{Comparison of different model architectures on the UCI HAR test set.}
    \label{tab:models1}
  \end{center}
\end{table}

\vspace{-0.3cm} 

Table \ref{tab:models1} shows that a model 196-16C\_4M\_1024D gives the best result, and we use it in our experiments. The model includes 196 filters of a  16x1 convolution layer followed by a 1x4 max pooling layer, then by 1024 units of a dense layer and finally a softmax layer.

\subsection{Settings for the experiments}

All experiments were written on Python 3 using the TensorFlow 2 library and run on a CPU Intel(R) Xeon(R) 2.30GHz (2 CPU cores, 12GB available RAM).

We used the same initial weights for all the models which were gotten by pre-training the chosen CNN (see Section 3.5) on a well-balanced small dataset (10 examples of each class).

We run $R=8$ communication rounds and $E=10$ epochs for the local training on a client side. We assumed that we have $K=5$ clients (one client generalize the influence of $K-1$ of them). The size of a dataset for each client $k$ and each round $r$ is $|D_{kr}| = 120$. We used a learning rate $\eta = 0.01$, dropout rate equal to 0.5, batch size $B = 32$ and SGD optimizer.  

\subsection{Demonstration of a problem}

\begin{figure}[!ht]
    % \centering
    % \subfloat[\centering \: \: CL]{{\includegraphics[width=2.62cm]{pics/cl0-cl.png} }}%
    % \:
    % \subfloat[\centering FCL ]{{\includegraphics[width=5.58cm]{pics/cl0-ft.png} }}%
    % \caption{Demonstration of catastrophic forgetting in standard CL (a) and FCL (b).}%
    \includegraphics[width=8.8cm]{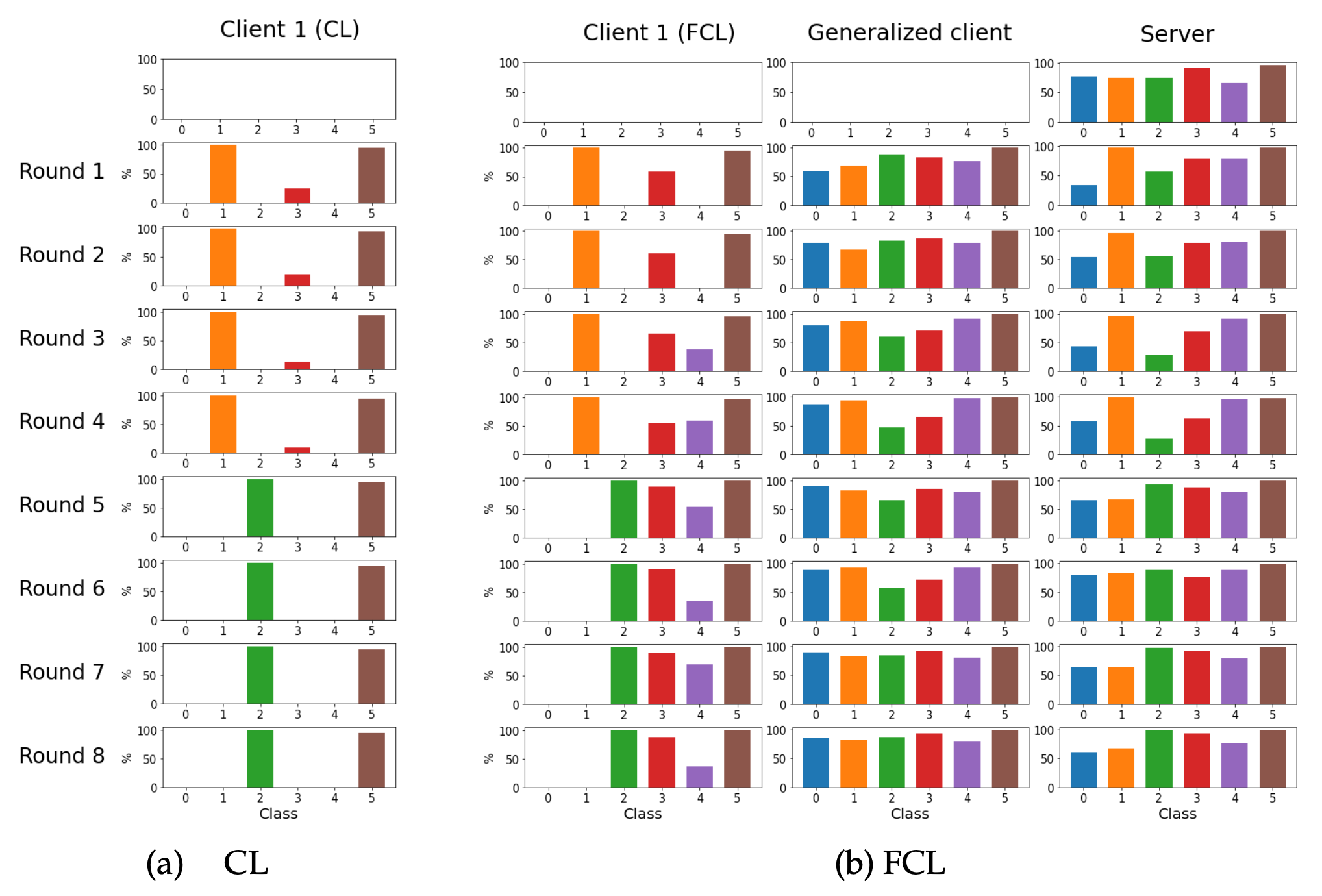}
    
    \vspace{-0.05cm}
    
    \caption{Demonstration of catastrophic forgetting in standard CL (a) and FCL (b).}%
    \label{fig:cat-for0}%
\end{figure}

In standard CL (Figure \ref{fig:cat-for0}(a)), we can see that class $C_1 = \{1\}$ was immediately forgotten when the learning task was changed on $C_2 = \{2\}$. But we can see always good performance on class 5 (laying), even if we don't have examples of these class during training. As we can see on Figure \ref{fig:pca} that the laying class is very distinguishable from the others, so the initial pre-trained model successfully learnt it, and model parameters used to classify it were not changed with further training. 

\smallskip

In FCL (Figure \ref{fig:cat-for0}(b)), we used simple Fine-Tuning of a model when we got a new data. We still can see immediate catastrophic forgetting of the task $1$ by client $1$. But Federated Learning successfully transferred to the client $1$ the knowledge about the other static actions (classes 3-5). 

%%%%%%%%%%%%%%%%%%%%%%%%%%%%%%%%%%%%%%%%%%%%%%%%%%%%%%%%%%
%%%%%%%%%%%%%%%%%%%%%%%%%%%%%%%%%%%%%%%%%%%%%%%%%%%%%%%%%%
%%%%%%%%%%%%%%%%%%%%%%%%%%%%%%%%%%%%%%%%%%%%%%%%%%%%%%%%%%

\newpage 

\section{Proposal}

To deal with the catastrophic forgetting in FCL, we propose a distillation-based approach inspired by \textit{Learning without Forgetting} (LwF) \cite{lwf} with the use of previous model on a client side and current server model as the teachers for the present client model.

\subsection{Baseline: FLwF}

Knowledge distillation was originally proposed to transfer the knowledge from a large model (teacher) to a smaller one (student) \cite{dist1}, \cite{dist2}. In LwF, authors use this technique in CL to prevent drifting while learning new tasks. For this, distillation loss was proposed to use while training the model.

First, we implemented a standard LwF method in FCL. In a standard LwF, there is one teacher model (past model of the client -- round $r-1$) and one student model (current client model -- round $r$). As we use an initial model pre-trained on all classes, each client has some knowledge about all $n=6$ classes from the beginning.

Output logits for the teacher classifier (past model of the client from a round $r-1$) is denoted in FCL as $\mathbf{o}^{r-1}(x)=\left[o^{r-1}_{1}(x), \ldots, o^{r-1}_{n}(x)\right]$, where  $x$ is an input to the network, and the output logits of the student classifier (current client model of a round $r$) is $\mathbf{o}^{r}(x)=\left[o^{r}_{1}(x), \ldots, o^{r}_{n}(x)\right]$.

The \textit{distillation loss} for the client $k$ and communication round $r$ for LwF approach in FCL is defined as:
\vspace{-0.2cm}
\begin{equation}
\label{eqn:dis_client}
L_{dis\_cl}(D_{kr}; \theta^{k}_{r}, \theta^{k}_{r-1}) = \sum_{x\in X_{kr}} \sum_{i=1}^{n}-\pi_{i}^{r-1}(x) \log \left[\pi_{i}^r(x)\right], \nonumber
\end{equation}
\vspace{-0.3cm}

 \noindent where $\theta^{k}_{r}$ is the weights of the current (student) model of the client $k$ in a communication round $r$ ($\theta^{k}_{r-1}$ -- previous (teacher) model),  $D_{kr} = \{X_{kr},Y_{kr}\}$ is the dataset used during  communication round $r$ by a client $k$, and $\pi_{i}^{r'}(x)$ are  temperature-scaled logits of the network: 

\vspace{-0.2cm}

\begin{equation}
\pi_{i}^{r'}(x) =\frac{e^{o_{i}^{r'}(x) / T}}{\sum_{j=1}^{n} e^{o_{j}^{r'}(x) / T}}, \nonumber
\end{equation}

\noindent where $T$ is the temperature scaling parameter \cite{dist2}. Temperature-scaled logits  $\pi_{i}^{r-1}(x) $ refer to predictions of the teacher model ($\mathbf{o}^{r-1}(x)$) and $\pi_{i}^{r}(x) $ refer to predictions of the student model ($\mathbf{o}^{r}(x)$).

A \textit{classification loss} (the softmax cross-entropy) in FCL is: 

\vspace{-0.4cm}
\begin{equation}
\label{eqn:loss_class}
L_{class}(D_{kr}; \theta^{k}_r) = \sum_{(x,y) \in D_{kr}} \sum_{i=1}^{n} -y_i \log \frac{exp(o^r_i(x)}{\sum_{j=1}^{n}exp(o^r_j(x))}, \nonumber
\end{equation}

\noindent where  $(x,y) \in D_{kr} = \{X_{kr}, Y_{kr}\}$ and $D_{kr} \subset D_{kr}^t$; $x$ is a vector of input features of a training sample, $y$ corresponds to some class of a set $C^t_k$ and presents as a one-hot ground truth label vector corresponding to $x$: $y\in\{0,1\}^{n=6}$.

The final loss in standard LwF for FCL for each client $k$ consists of a classification loss and distillation loss computed with the current model of client $k$ for round $r$  and previous model of the client $k$ for round $r-1$: 

\vspace{-0.15cm}
\begin{equation}
L_{FLwF} = \alpha L_{class} + (1-\alpha)L_{dis\_cl}, \nonumber
\end{equation}

\noindent where $\alpha$ is a scalar which regularizes influence of each term.

\smallskip

We name this implementation as \textbf{Federated Learning without Forgetting (FLwF)}.

\subsection{Proposal: FLwF-2T}

We propose to take an advantage of the server which keeps a general knowledge about all the clients. Our goal is to use the server as a second teacher and to send its knowledge to a student client model. The first teacher (past model of a client) can increase the specificity performance of a student (client) which allows to be still good on tasks it has learned before. The second teacher (server) can improve general features of a client model by transferring the knowledge from all the other clients and avoid over-fitting of a client model on its new task. 

\smallskip

We denote $\hat{\mathbf{o}}^{r-1}(x)=\left[\hat{o}_{1}^{r-1}(x), \ldots, \hat{o}_{n}^{r-1}(x)\right]$ as the output (before the activation) of the global model (server network) which was gotten after the aggregation of clients' models trained after the round $r-1$. The temperature-scaled logits of the server network are defined as follows: 

\vspace{-0.15cm}
\begin{equation}
\hat{\pi}_{i}^{r-1}(x) =\frac{e^{\hat{o}^{r-1}_{i}(x) / T}}{\sum_{j=1}^{n} e^{\hat{o}^{r-1}_{j}(x) / T}}. \nonumber
\end{equation}
\vspace{-0.05cm}

The \textit{distillation loss for the server model} is defined as:

\vspace{-0.5cm}
\begin{equation}
\label{eqn:dis_server}
L_{dis\_serv}(D_{kr}; \theta^{k}_{r}, \theta_{r-1}) = \sum_{x\in X_{kr}} \sum_{k=1}^{n}-\hat{\pi}_{k}^{r-1}(x) \log \left[\pi_{k}^r(x)\right], \nonumber
\end{equation}

\vspace{-0.2cm}

\noindent where  $\theta^{k}_{r}$ is weights of the current (student) model of the client $k$ in a communication round $r$ and  $\theta_{r-1}$ is weights of a global model on a server gotten by aggregation of clients' models after communication round $r-1$.

The final loss for the proposed method is:

\vspace{-0.4cm}
\begin{equation}
\label{eqn:final}
L_{FLwF-2T} = \alpha L_{class} + \beta L_{dis\_cl} + (1-\alpha -\beta) *  L_{dis\_serv}, \nonumber
\end{equation}

\noindent where $\alpha$ and $\beta$ are scalars which regularize influence of terms. $L_{dis\_cl}$ refers to the distillation loss from past model (Teacher 1), defined in a Section 4.1, and $L_{dis\_serv}$ refers to a server model (Teacher 2). 

For the first communication round, the model uses only the server model as a teacher.

The proposed solution helps to transfer the knowledge from the server  and decrease the forgetting of previously learnt tasks.  It is a regularization-based approach as it prevents activation drift while learning new tasks. We name it \textbf{Learning without Forgetting - 2 Teachers (LwF-2T)}. The pseudo-code of FLwF-2T is presented in Algorithm \ref{alg:fcl}, where loss function $L_{FLwF-2T}(... ; b)$ is counted on a batch $b$.  %cost function
%counted on a batch $b$
%where $B$ is a local mini-batch size, $\eta$ is a learning rate for the gradient step in model updating, $L_{FLwF-2T}(... ; b)$  is a loss function %cost function
%counted on a batch $b$.

\begin{algorithm}[h!]
		\caption{FLwF-2T.}
		\label{alg:fcl}
		\begin{algorithmic}[1]
            \Procedure{Server executes:}{}
			\State 
			initialize server model by $\theta_0$
			\For {round  $r = 1,2,...,R$} 
			\State m = 0
			\For {client $k = 1, ..., K$ \textbf{in parallel}} 
			
			\State $\theta_{r}^k \leftarrow$ ClientUpdate$( k, r, \theta_{r-1},\theta_{r-1}^k )$ 
			
			\State $m_k = |D_{kr}|$ \hspace{10mm} // \textit{Size of a dataset} $D_{kr}$
			\State $m += m_k$
			
			\EndFor
			\State  $\theta_{r} \leftarrow \sum\limits_{k=1}^{K} \frac{m_{k}}{m} \theta_{r}^{k}$
			\EndFor
			
			\EndProcedure
			
			\smallskip

            \Function{ClientUpdate}{$k, r, \theta_{r-1},\theta_{r-1}^k$}:    
			
			\State $\mathcal{B} \leftarrow\left(\right.$split $D_{kr}$ into batches of size $\left.B\right)$
			\State $\theta^k_r = \theta_{r-1} $
			\For {each local epoch $i$ from $1$ to $E$ }
	        \For {batch $b \in \mathcal{B} $}
	        \State $\theta^k_r \leftarrow \theta^k_r-\eta \nabla L_{FLwF-2T}(\theta, \theta_{r-1},\theta_{r-1}^k; b)$
	        \EndFor
	        \EndFor
	        \State return $\theta^k_r$ to the server
			\EndFunction
		\end{algorithmic}
\end{algorithm}

\vspace{-0.35cm}

\subsection{Metrics}

To evaluate models performance, we use several metrics divided into two groups. In a first group, we use the metrics which describe FL side of a process: generality of models and performance on a specific knowledge of a client. Second group of metrics evaluate a forgetting of the learnt tasks from a CL side of a process. 
Let's define $a_{t,d}^{kr}$ as an accuracy of the model trained during communication round $r$ on a task $d$ after learning task $t$ ($d \leqslant t$) for the client $k$. To compute $a_{t,d}^{kr}$ we take all the examples of classes corresponding to a task $d$ from the test set, calculate an accuracy of the model, which was trained during communication round $r$, on them after the learning task $t$. An accuracy $a_{0}^{kr}$ is calculated on the whole test set after communication round $r$ for the client $k$ or the server.

\smallskip

\textbf{\large FL metrics:}

\begin{itemize}
    \vspace{-0.05cm}\item To evaluate generality of a model, we calculate a \textbf{general accuracy} ($A_{gen}^k$):
    
    \vspace{-0.55cm}
    \begin{equation}
    \label{eq:gen}
    A_{gen}^k = \frac{1}{R} \sum_{r = 1}^R a_{0}^{kr}. \nonumber
    \end{equation}
    \vspace{-0.25cm}

We compute this general accuracy for the specified client, generalized client and server.

    \vspace{-0.06cm}\item To evaluate a performance of a model on a specific
    knowledge of a client, we calculate a \textbf{personal accuracy} ($A_{per}^k$):
    \vspace{-0.29cm}
    \begin{equation}
    \label{eq:per}
    A_{per}^k = \frac{1}{R} \sum_{r = 1}^R a_{per}^{kr}, \nonumber
\end{equation}
\vspace{-0.25cm}

\noindent where an accuracy  $a_{per}^{kr}$ is calculated on classes which  were already learnt by a client $k$ during the rounds $1,...,r$. We compute the personal accuracy only for  for the clients with specific CL scenario.

\end{itemize}

\smallskip

\textbf{\large CL metrics:}

To evaluate how a model forgets previously learnt tasks, we use an average accuracy at task $t$ on all learnt before tasks and forgetting \cite{class-inc}. We compute them only for the clients with specific CL scenario.

\begin{itemize}
    \vspace{-0.1cm}\item \textbf{Average accuracy at task $t$} ($A_t^k$) calculated on all learnt before tasks  for the client $k$:
    \vspace{-0.25cm}
    \begin{equation}
    \label{eq:avgacc}
    A_{t}^k = \frac{1}{t} \sum_{d = 1}^{t} a_{t,d}^{k}, \hspace{7mm} a_{t,d}^{k} = a_{t,d}^{k} = \frac{1}{r_k^t} \sum_{r'= (r_k^{t})'}^{(r_k^{t})' + r_k^t} a_{t,d}^{kr'}, \nonumber
\end{equation}
\vspace{-0.3cm}

where $(r_k^{t})' = \sum_{t=1}^{t-1}r^{t}_{k}$.

\item \textbf{Forgetting} ($f_{t,d}^k$) shows how the model forgets the knowledge about task $d$ after the learning a task $t$ for the client $k$ :
    \vspace{-0.2cm}
    \begin{equation}
    \label{eq:forgetting}
    f_{t,d}^{k}= \max _{i \in\{d, \ldots, t-1\}}\{a_{i,d}^{k}\}-a_{t,d}^{k}. \nonumber
\end{equation}
\vspace{-0.3cm}

It can be averaged as  $F_{t}^k=\frac{1}{t-1} \sum_{d=1}^{t-1} f_{t,d}^{k}$. The higher $F_{t}^k$, the more a model forgets.

\end{itemize}

%%%%%%%%%%%%%%%%%%%%%%%%%%%%%%%%%%%%%%%%%%%%%%%%%%%%%%%%%%
%%%%%%%%%%%%%%%%%%%%%%%%%%%%%%%%%%%%%%%%%%%%%%%%%%%%%%%%%%
%%%%%%%%%%%%%%%%%%%%%%%%%%%%%%%%%%%%%%%%%%%%%%%%%%%%%%%%%%
%%%%%%%%%%%%%%%%%%%%%%%%%%%%%%%%%%%%%%%%%%%%%%%%%%%%%%%%%%
%%%%%%%%%%%%%%%%%%%%%%%%%%%%%%%%%%%%%%%%%%%%%%%%%%%%%%%%%%
%%%%%%%%%%%%%%%%%%%%%%%%%%%%%%%%%%%%%%%%%%%%%%%%%%%%%%%%%%
%%%%%%%%%%%%%%%%%%%%%%%%%%%%%%%%%%%%%%%%%%%%%%%%%%%%%%%%%%
%%%%%%%%%%%%%%%%%%%%%%%%%%%%%%%%%%%%%%%%%%%%%%%%%%%%%%%%%%
%%%%%%%%%%%%%%%%%%%%%%%%%%%%%%%%%%%%%%%%%%%%%%%%%%%%%%%%%%

\section{Experiments}

In this section, we present experiments where we compared different methods in FCL. We also propose to use different strategies of training for generalized client and use exemplars from learnt before tasks. 

\vspace{-0.12cm}
\subsection{Comparison of methods}

\vspace{-0.05cm}

We compared following methods in FCL: Fine-Tuning in simple Continual Learning (CL-FT), Fine-Tuning in Federated Continual Learning (FCL-FT), Federated Learning without Forgetting (FLwF), our proposed method with two teachers (FLwF-2T) and Offline Learning on all training data for each client (all data). Results in a first and second parts of Tables \ref{tab:fl_main}, \ref{tab:cl_main}.

For the methods FLwF and FLwF2, we used a temperature  scaling  parameter $T=2$ as commonly used in other experiments \cite{bic}, \cite{eeil}, \cite{class-inc}. By using grid search we found that for FLwF $\alpha = 0.001$ and for FLwF  $\alpha = 0.001$, $\beta = 0.7$ shows the best performance.

\vspace{-0.07cm}
\subsection{Generalized client: Fine-Tuning}

\vspace{-0.05cm}
In this experiment, we propose to choose a strategy of training depending on a current local data. If the local data during current communication rounds is well-balanced and contains all classes, we propose to use simple Fine-Tuning not to be dependant on a server global model which is influenced by other clients, if not, we offer to implement CL approaches to remember our past and to transfer as more available knowledge from the server as possible.

We define the strategy when Client 1 uses FLwF (FLwF-2T) and Generalized client uses FT as FLwF/FT (FLwF-2T/FT). Results in a fourth part of Tables \ref{tab:fl_main}, \ref{tab:cl_main}.

\vspace{-0.2cm}
\begin{table*}[hbt!]
\parbox{.54\linewidth}{
\centering
\begin{tabular}{|c"c|c|c"c|}
      \hline
      \textbf{Method} & \textbf{$A_{gen}^1$} & \textbf{$A_{gen}^g$} & \textbf{$A_{gen}^{server}$} & \textbf{$A_{per}^1$} \\ [0.1cm]
    
      \hline
      CL-FT & 0.338 & 0.800 & -  & 0.750\\
      all data & 0.715 & 0.899 & 0.865 & 0.977\\ [0.03cm]
      \hline 
      FCL-FT & 0.478 & \textbf{0.794} & \textbf{0.714} & \textbf{0.750}\\
      FLwF & 0.629 & 0.673 & 0.671 & 0.628 \\
      \textbf{FLwF-2T} &  \textbf{0.655} & 0.679 &  0.680 & 0.629\\ [0.05cm]
      \hline 
      FCL-FT + ex &  0.622 & 0.675 &  0.672 & 0.666 \\
      FLwF + ex & 0.633 & 0.674 & 0.675 & 0.659 \\
      \textbf{FLwF-2T + ex} &  \textbf{0.658} & \textbf{0.683} &  \textbf{0.681} & \textbf{0.664} \\
      \hline
      FLwF/FT & 0.708 & 0.783 & 0.781 & 0.755 \\
      \textbf{FLwF-2T/FT} &  \textbf{0.753} & \textbf{0.802} &  \textbf{0.797} & \textbf{0.798} \\ [0.05cm]
      \hline
      FLwF/FT + ex &  0.705 & \textbf{0.789} &  \textbf{0.779} & 0.748 \\
      FLwF-2T/FT + ex & \textbf{0.750} & 0.781 & 0.778 & \textbf{0.755} \\
      \hline
\end{tabular}
\caption{\textbf{FL metrics}: general accuracy ($A_{gen}^k$) and personal accuracy ($A_{per}^k$) for Client 1 ($k=1$), Generalized client ($k=g$) and Server ($k=server$).}
\label{tab:fl_main}
}
\hfill
\parbox{.42\linewidth}{
\centering
\begin{tabular}{|c"c"c|}
      \hline
      \textbf{Method} & \textbf{$A^1_2$} & \textbf{$F_2^1$ } \\ [0.1cm]

      \hline
      CL-FT & 0.5 & 1  \\
      all data & 0.981 & -0.007 \\
      \hline 
      FCL-FT & 0.5 & 1 \\
      FLwF & 0.535 & 0.595 \\
      \textbf{FLwF-2T} &  \textbf{0.578} & \textbf{0.418} \\
      \hline
      FCL-FT + ex & 0.548 & 0.617  \\
      FLwF + ex & 0.57 & 0.53 \\
      \textbf{FLwF-2T + ex} &  \textbf{0.622} & \textbf{0.29} \\
      \hline
      FLwF/FT & 0.696 & 0.392 \\
      \textbf{FLwF-2T/FT} & \textbf{0.76} & \textbf{0.212}\\
      \hline
      FLwF/FT + ex &  0.678 & 0.407\\
      \textbf{FLwF-2T/FT + ex} & \textbf{0.746} & \textbf{0.12}  \\
      \hline
\end{tabular}
\caption{\textbf{CL metrics}: average accuracy at task $t$ ($A^k_t$) and forgetting ($F_t^k$) for Client 1 ($k=1$) and task 2 ($t=2$). }
\label{tab:cl_main}
}
\end{table*}

\vspace{0.02cm}
\subsection{Saving the exemplars}

\vspace{-0.05cm}

\hspace{4mm} In this experiment, we saved the exemplars of learned task in a memory. For each round, we realized the following procedure: if the task is new, we save 10 examples which are corresponded to this task in our memory; if the model has learnt this task already, we refresh out memory stock with new examples which are corresponded to this task. We implemented random sampling strategy as it shows competitive results compare to another strategies, used in class-incremental CL, and don't require much computational resources \cite{class-inc}. Results in a third and fifth part of Tables \ref{tab:fl_main}, \ref{tab:cl_main}.

\vspace{-0.15cm}
\subsection{Discussion of results}

\vspace{-0.05cm}

\begin{figure}[!ht]
\centering
\includegraphics[width=0.45\textwidth]{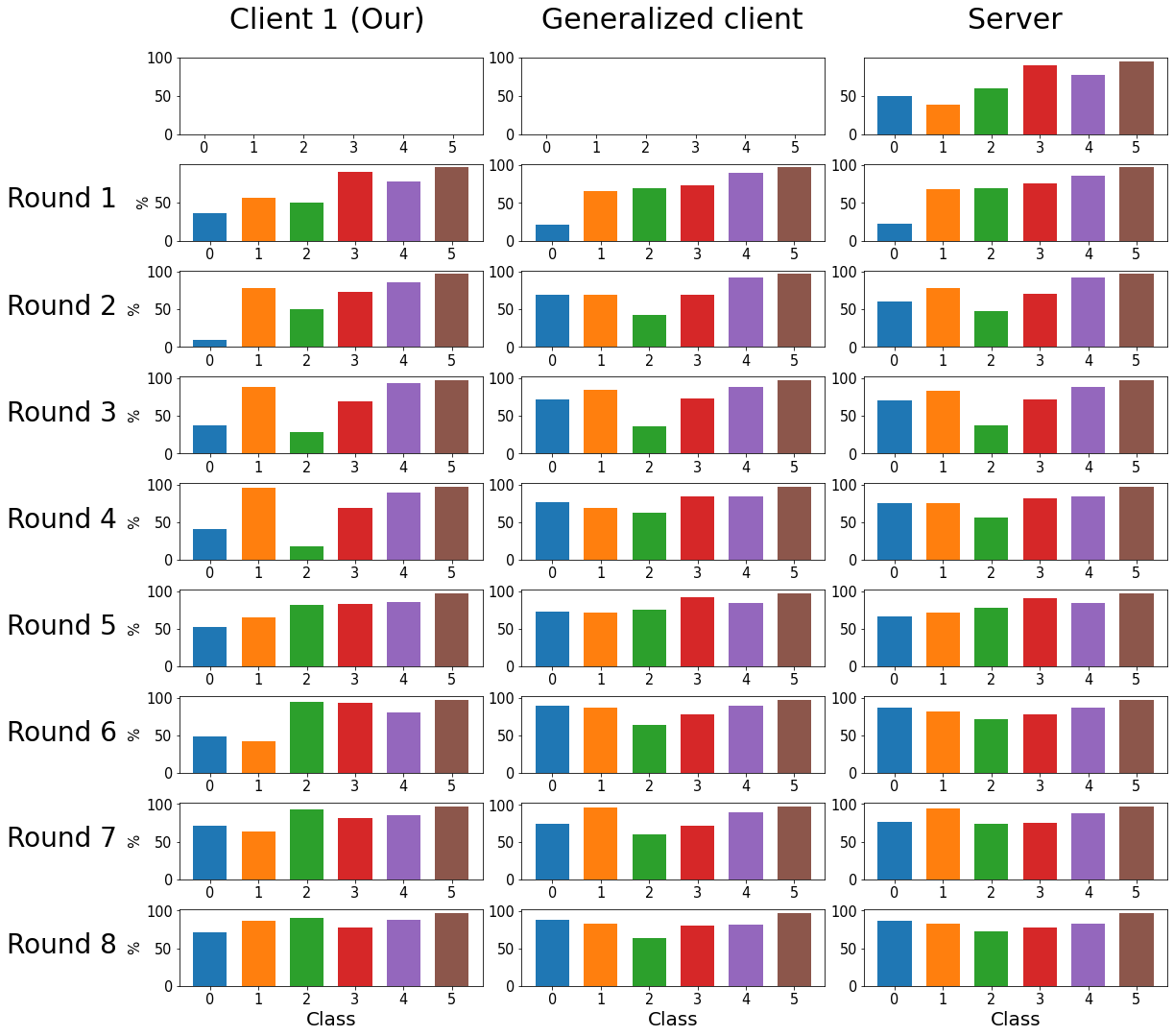}
\caption{The best gotten result according to Tables \ref{tab:fl_main}, \ref{tab:cl_main} shown by FLwF-2T/FT (our).}
\label{fig:fl2}
\end{figure}

If we compare Figure \ref{fig:cat-for0}(a) and Figure \ref{fig:fl2}(b), we will see that our Continual Learning approach (FLwF-2T) helps both to increase the generality of models and to keep the knowledge from learned tasks.

First, we compared different methods in FCL for Client 1 and its specific CL scenario.  \textbf{For Client 1, our method FLwF-2T shows the best general accuracy among all the methods } (Table \ref{tab:fl_main}). It also outperforms FLwF on a Generalized client and on the Server. Personal accuracy of FLwF-2T is decreased compared to FCL-FT due to the significant over-fitting while FT. 
We also can see that FL with implemented CL method helps to deal with catastrophic forgetting. And our method FLwF-2T shows the best result among others in dealing with catastrophic forgetting for Client 1 (Table \ref{tab:cl_main}).

Then, we proposed to use different strategies of training depending on a local data for the current communication round.  We can see that among compared methods, our approach with the use of FLwF-2T on a Client 1 and FT on a  Generalized client significantly outperforms all other methods both in capturing generality of a model and keeping the past knowledge. So, \textbf{implementing different strategies of training  for different clients depending on a local data can improve the general performance in FCL.}

Then, we made experiments with the saving of exemplars from learnt tasks. We can see, that \textbf{adding exemplars allowed to increase performance for both CL metrics for all the observed methods}. When we don't choose the training strategy for the clients, FLwF-2T + ex shows the best result among all the methods with adding exemplars. When generalized client is trained by FT,  FLwF/FT + ex shows better generalization accuracy for the generalized client and the server than FLwF-2T/FT + ex.

FLwF-2T/FT -- our proposed method -- shows the best result among all the observed methods for both  FL and CL metrics according to Tables \ref{tab:fl_main}, \ref{tab:cl_main} (Figure \ref{fig:fl2}).

\vspace{-0.3cm}
\vspace{-0.75cm}

\section{Conclusion}

\vspace{-0.09cm}

Our study presents a novel framework for the fusion of two learning paradigms: Federated Learning and Class-Incremental learning with the application in Pervasive Computing (HAR on mobile devices). 

We showed that without CL approaches in FL a rapid forgetting of learnt tasks is also happened. We saw that the close relation between classes in a UCI HAR dataset influence a faster forgetting of some of them. Finally, we proposed distillation-based approach FLwF-2T for FCL, which doesn't require high computational and storage resources which is crucial for mobile devices as it uses for the training only a past model of the client and a global model sent by the server which would anyway available for the client during each communication round. We showed that it allows to increase a general knowledge for a client with specific CL scenario and to keep it's private past knowledge. 

\bibliographystyle{plain}
%\bibliography{ijcai21}

{\footnotesize
\bibliography{ijcai21}}

\begin{thebibliography}{10}

\bibitem{mas}
R.~Aljundi, F.~Babiloni, M.~Elhoseiny, M.~Rohrbach, and T.~Tuytelaars.
\newblock Memory aware synapses: Learning what (not) to forget.
\newblock {\em European Conference on Computer Vision}, 2018.

\bibitem{uci}
Davide Anguita, Alessandro Ghio, Luca Oneto, Xavier Parra, and Jorge~L.
  Reyes-Ortiz.
\newblock A public domain dataset for human activity recognition using
  smartphones.
\newblock In {\em 21th European Symposium on Artificial Neural Networks,
  Computational Intelligence and Machine Learning, ESANN 2013}, pages 24--26,
  2013.

\bibitem{fedper}
Manoj~Ghuhan Arivazhagan, Vinay Aggarwal, Aaditya~Kumar Singh, and Sunav
  Choudhary.
\newblock Federated learning with personalization layers.
\newblock {\em arXiv:1912.00818v1}, 2019.

\bibitem{mcmahan}
McMahan~H. B., Moore E., Ramage D., and Hampson S.
\newblock Communication-efficient learning of deep networks from decentralized
  data.
\newblock {\em International Conference on Artificial Intelligence and
  Statistics (AISTATS)}, 2017.

\bibitem{becker}
C.~Becker, C.~Julien, P.~Lalanda, and F.~Zambonelli.
\newblock Pervasive computing middleware: current trends and emerging
  challenges.
\newblock {\em CCF Transactions on Pervasive Computing and Interaction, vol.
  1}, 2019.

\bibitem{il2m}
E.~Belouadah and A.~Popescu.
\newblock Il2m: Class incremental learning with dual memory.
\newblock {\em International Conference on Computer Vision,}, 2019.

\bibitem{blachon:hal-01082580}
David Blachon, Doruk Cokun, and Fran{\c c}ois Portet.
\newblock {On-line Context Aware Physical Activity Recognition from the
  Accelerometer and Audio Sensors of Smartphones}.
\newblock In {\em {European Conference on Ambient Intelligence}}, volume 8850
  of {\em Ambient Intelligence}, pages 205--220, Eindhoven, Netherlands, 2014.

\bibitem{dist1}
C.~Bucilua, R.~Caruana, and A.~Niculescu-Mizil.
\newblock Model compression.
\newblock {\em International Conference on Knowledge Discovery and Data
  Mining}, 2006.

\bibitem{fcl-nodnn}
Fernando~E. Casado, Dylan Lema, Roberto Iglesias, Carlos~V. Regueiro, and Sene
  n~Barroa.
\newblock Federated and continual learning for classification tasks in a
  society of devices.
\newblock {\em arXiv:2006.07129v2 [cs.LG]}, 2021.

\bibitem{eeil}
Francisco~M. Castro, Manuel~J. Mar{\'{\i}}n{-}Jim{\'{e}}nez, Nicol{\'{a}}s
  Guil, Cordelia Schmid, and Karteek Alahari.
\newblock End-to-end incremental learning.
\newblock {\em European Conference on Computer Vision}, 2018.

\bibitem{percom}
Sannara Ek, Francois Portet, Philppe Lalanda, and German Vega.
\newblock A federated learning aggregation algorithm for pervasive computing:
  Evaluation and comparison.
\newblock In {\em Proceedings of the 20th International Conference on Pervasive
  Computing and Communications (PerCom 2022)}, 2021.

\bibitem{bonawitx}
Keith~Bonawitz et~al.
\newblock Towards federated learning at scale: system design.
\newblock In {\em Proceedings of the 2nd SysML Conference}, Palo Alto, CA, USA,
  2019.

\bibitem{pathint}
F.Zenke, B.Poole, and S.Ganguli.
\newblock Continual learning through synaptic intelligence.
\newblock {\em International Conference on Machine Learning}, 2017.

\bibitem{dist2}
G.~Hinton, O.~Vinyals, and J.~Dean.
\newblock Distilling the knowledge in a neural network.
\newblock {\em NIPS Deep Learning and Representation Learning Workshop}, 2015.

\bibitem{lucir}
S.~Hou, X.~Pan, C.~C. Loy, Z.~Wang, and D.~Lin.
\newblock Learning a unified classifier incrementally via rebalancing.
\newblock {\em International Conference on Computer Vision,}, 2019.

\bibitem{ignatov}
Andrey Ignatov.
\newblock Real-time human activity recognition from accelerometer data using
  convolutional neural networks.
\newblock {\em Applied Soft Computing}, pages 915--922, 2018.

\bibitem{pca}
Ian~T. Jolliffe and Jorge Cadima.
\newblock Principal component analysis: a review and recent developments.
\newblock {\em Phil. Trans. R. Soc.}, 2016.

\bibitem{ewc}
J.~Kirkpatrick, R.~Pascanu, N.~Rabinowitz, J.~Veness, G.~Desjardins, A.~A.
  Rusu, K.~Milan, J.~Quan, T.~Ramalho, and A.~Grabska-Barwinska et~al.
\newblock Overcoming catastrophic forgetting in neural networks.
\newblock {\em National Academy of Sciences, vol. 114, no. 13}, 2017.

\bibitem{task-inc}
Matthias~De Lange, Rahaf Aljundi, Marc Masana, Sarah Parisot, Xu~Jia, Alesˇ
  Leonardis, Gregory Slabaugh, and Tinne Tuytelaars.
\newblock A continual learning survey: Defying forgetting in classification
  tasks.
\newblock {\em arXiv:1909.08383v2}, 2020.

\bibitem{lwf}
Z.~Li and D.~Hoiem.
\newblock Learning without forgetting.
\newblock {\em IEEE Transactions on Pattern Analysis and Machine Intelligence,
  vol. 40}, 2017.

\bibitem{8}
W.~Y.~B. Lim, N.~C. Luong, D.~T. Hoang, Y.~Jiao, Y.~C. Liang, Q.~Yang,
  D.~Niyato, and C.~Miao.
\newblock Federated learning in mobile edge networks: A comprehensive survey.
\newblock {\em IEEE Communications Surveys Tutorials, vol. 22, no. 3, pp.
  2031–2063}, 2020.

\bibitem{class-inc}
Marc Masana, Xialei Liu, Bartłomiej Twardowski, Mikel Menta, Andrew~D.
  Bagdanov, and Joost van~de Weijer~C.
\newblock Class-incremental learning: survey and performance evaluation.
\newblock {\em arXiv:2010.15277v1}, 2020.

\bibitem{fedavg}
H~Brendan McMahan, Eider Moore, Daniel Ramage, Seth Hampson, and et~al.
\newblock Communication-efficient learn- ing of deep networks from
  decentralized data.
\newblock {\em arXiv preprint arXiv:1602.05629}, 2016.

\bibitem{catastrophic-forgetting}
McCloskey Michael and Cohen~Neal J.
\newblock Catastrophic interference in connectionist networks: The sequential
  learning problem. psychology of learning and motivation.
\newblock {\em 24. pp. 109–165. doi:10.1016/S0079-7421(08)60536-8. ISBN
  978-0-12-543324-2.}, 1989.

\bibitem{lara}
O.D.Lara and M.A.Labrador.
\newblock A mobile platform for real-time human activity recognition.
\newblock {\em IEEE Consumer Communications and Networking Conference (CCNC)},
  pages 667--671, 2012.

\bibitem{icarl}
S.-A. Rebuffi, A.~Kolesnikov, G.~Sperl, and C.~H. Lampert.
\newblock icarl: Incremental classifier and representation learning.
\newblock {\em Conference on Computer Vision and Pattern Recognition}, 2017.

\bibitem{cl}
Sebastian Thrun.
\newblock - {A} {Lifelong} {Learning} {Perspective} for {Mobile} {Robot}
  {Control}.
\newblock In Volker Graefe, editor, {\em Intelligent {Robots} and {Systems}},
  pages 201--214. Elsevier Science B.V., Amsterdam, 1995.

\bibitem{fedma}
H.~Wang, M.~Yurochkin, Y.~Sun, D.~Papailiopoulos, and Y.~Khazaeni.
\newblock Federated learning with matched averaging.
\newblock {\em International Conference on Learning Representations}, 2020.

\bibitem{wu-cnn}
Q.~Wu, K.~He, and X.~Chen.
\newblock Personalized federated learning for intelligent iot applications: A
  cloud-edge based framework.
\newblock {\em IEEE Open Journal of the Computer Society}, 2020.

\bibitem{bic}
Y.~Wu, Y.~Chen, L.~Wang, Y.~Ye, Z.~Liu, Y.~Guo, and Y.~Fu.
\newblock Large scale incremental learning.
\newblock {\em International Conference on Computer Vision}, 2019.

\bibitem{fcl}
Xin Yao and Lifeng Sun.
\newblock Continual local training for better initialization of federated
  models.
\newblock {\em arXiv:2005.12657v1 [cs.LG]}, 26 May 2020.

\bibitem{fedweit}
Jaehong Yoon, Wonyong Jeong, Giwoong Lee, Eunho Yang, and Sung~Ju Hwang.
\newblock Federated continual learning with weighted inter-client transfer.
\newblock {\em arXiv:2003.03196 [cs.LG]}, 2020.

\end{thebibliography}

\end{document}